\documentclass{article}

\usepackage{arxiv}
\usepackage{algorithm}  
\usepackage[noend]{algpseudocode}  
\usepackage[latin1]{inputenc} 
\usepackage{hyperref}   
\usepackage{url}            
\usepackage{booktabs}       
\usepackage{nicefrac}       
\usepackage{microtype}      
\usepackage{amsmath,amssymb,amsfonts}
\usepackage{graphicx}
\usepackage{textcomp}
\usepackage{xcolor}
\usepackage{subcaption}
\usepackage{booktabs}
\usepackage{mathrsfs}
\usepackage[square,numbers,sort&compress]{natbib}

\DeclareMathOperator{\ODESOLVE}{\textit{ODESolve}}
\DeclareMathOperator{\tuple}{\textit{tuple}}

\renewcommand{\Re}{\mathbb{R}}

\def\BibTeX{{\rm B\kern-.05em{\sc i\kern-.025em b}\kern-.08em
    T\kern-.1667em\lower.7ex\hbox{E}\kern-.125emX}}

\title{Neural ODEs for Informative Missingess in Multivariate Time Series}

\author{ \href{https://orcid.org/0000-0001-9051-1370}{\includegraphics[scale=0.06]{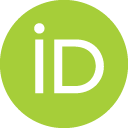}\hspace{1mm}Mansura Habiba} \\
	Cloud Solutions Architect\\
	IBM in Ireland\\
	Dublin, Ireland \\
	\texttt{mansura.habiba@gmail.com} \\
	\And
	\href{https://orcid.org/0000-0003-0521-4553}{\includegraphics[scale=0.06]{orcid.png}\hspace{1mm}Barak A. Pearlmutter} \\
	Department of Computer Science \& Hamilton Institute\\
	Maynooth University\\
	Maynooth, Ireland \\
	\texttt{barak@pearlmutter.net} \\
}

\hypersetup{
pdftitle={Neural ODEs for Informative Missingess in Multivariate Time Series},
pdfsubject={q-bio.NC, q-bio.QM},
pdfauthor={Mansura Habiba, Barak A. Pearlmutter},
pdfkeywords={ GRU-D, Informative missingness, Neural Ordinary Differentiation Equations},
}

\begin{document}
\maketitle

\begin{abstract}
Informative missingness is unavoidable in the digital processing of continuous time series, where the value for one or more observations at different time points are missing. Such missing observations are one of the major limitations of time series processing using deep learning.  Practical applications, e.g., sensor data, healthcare, weather, generates data that is in truth continuous in time, and informative missingness is a common phenomenon in these datasets. These datasets often consist of multiple variables, and often there are missing values for one or many of these variables. This characteristic makes time series prediction more challenging, and the impact of missing input observations on the accuracy of the final output can be significant. A recent novel deep learning model called GRU-D is one early attempt to address informative missingness in time series data. On the other hand, a new family of neural networks called Neural ODEs (Ordinary Differential Equations) are natural and efficient for processing time series data which is continuous in time. In this paper, a deep learning model is proposed that leverages the effective imputation of GRU-D, and the temporal continuity of Neural ODEs. A time series classification task performed on the PhysioNet dataset demonstrates the performance of this architecture.
\end{abstract}

\keywords{ GRU-D\and Informative missingness \and Neural Ordinary Differentiation Equations}

\section{Introduction}
\label{sec:intro}
Continuous time series usually consists of a series of data collected on different time points at a different sampling frequencies. Based on the sampling frequency as well as the availability of data, some time points in the series can have missing observations for many of the time point in a continuous time series. These missing values can influence the result of different time series problems, e.g. classification, prediction using deep learning. \cite{che2018recurrent} demonstrates some findings using the MIMIC-III datasets \cite{johnson2016mimic} which reflects the significance of missing patterns in time series prediction tasks. If the value for certain variables in a multivariate time series is missing for a significant time, the impact of corresponding input variables fades away over time and can result in inaccurate result. Over time various experiments are being used to process data with missing value in order to overcome the imposed challenges. These techniques are suitable for small as well as a simple time series. As the complexity, dynamics, length, sampling frequencies, number of the variable of the time series increases, these techniques become irrelevant. Some of the methods are as following
\begin{itemize}
\item Omit the missing data and perform the task on only available observations of the time series. This approach helps to ignore the unavailable data. However, if the missing rate is high and a significant amount of data is ignored, this solution often results in an inaccurate outcome and misguiding prediction.
\item Use data imputation to fill out the missing observation with substitutes value. There are several imputation techniques to determine substitute value. The limitation of this approach is that its only suitable for simple time series. It is often too hard to find right substitute values for complex time series. For complex time series, rather than using a single data imputation method, multiple imputation methods are often used combinedly to deal with the complexity of data series as well as to reduce the uncertainty.
\item Another practice is to apply data imputation multiple times iteratively with a target to reach an average value for the missing observations. However, data imputation only effective for simple data series with fixed missing rate and fixed time series length. In reality, time series are often variable in length and missing observations occur in completely random order.
\end{itemize}

Among all different Neural network families, Recurrent Neural Networks (RNN) have shown significant efficiency in case of solving missing pattern in time series with different gating units as well as their capacity of storing memories. These days, various time series tasks such as classification, prediction, generation are usually solved using RNN models. Different gating units for RNNs (e.g., Long Short-Term Memory (LSTM) \cite{hochreiter1997long}, Gated Recurrent Units (GRU) \cite{cho2014learning}, and GRU-D \cite{che2018recurrent}) typically consider time series as a dynamical system of the discrete and fixed time step. Theoretically, fixed step can be enough for time series modelling if it is too small. However, real-world time series data are usually sampled at an irregular rate. For some applications such as sensors, a short time step is necessary to cope with the higher data sampling frequencies. On the other hand, patients health record needs to be sampled at a higher time step, as the time gap between two consecutive visits of a patient can be very long. Due to the irregular data sampling rate and variable length, multivariate time series are very complex in nature. Therefore, it requires a structured and dynamic model to defeat the uncertainty of informative missingness.

\cite{chen2018neural,rubanova2019latent} try to handle missing pattern in continuous time series with data imputation. GRU-D \cite{che2018recurrent} successfully exploits the strength of RNN models for time series to capture the long term dependencies in multivariate time series.  Another recent family of neural networks, Neural Ordinary Differential Equations (ODE-NN) \cite{chen2018neural, BRYSON62, PEARLMUTTER89A} helps to solve time series by using black-box differential equation solver. This kind of neural network leverages the initial value problem to compute the hidden dynamics ($f$) as a function of continuous time at any time $t$. As shown in \eqref{eq:ht}, ODE-NN can compute the hidden state of time $t$ ($h_{t} $) from the initial hidden state($ h_{0} $) and the hyperparameters ($ \theta_{t-1} $) update over time. Here $t \in \{0,\ldots,T\}$  and $h_{t} \in \Re$.

\begin{equation}
\label{eq:ht}
h_{t} = h_{t-1} + f(h_{0}, \theta_{t-1})
\end{equation}

In this paper, two different neural network models are introduced. Both models leverage the differential equation solver in order to compute the hidden dynamics of the model. They uses differential equation solver to impute data, where the missing observations of variables are replaced by the derivative of the value of available observations of corresponding variables. Over time the decay in hidden dynamics, as well as input, has a significant impact on the final output on multivariate time series. This work computes the decay rate as the derivatives of time ($t$), therefore, the decay rate can control the gradient optimization of the model over time. First model use to generate the hidden dynamics of GRU-D model as continuous time dynamics using differential equation solver. Second model compute decay rate in addition to continuous hidden dynamics of GRU-D model. In this work, the time series are considered a function of time $t$, rather than discrete sequence. The second model proposed in this paper,shows an efficient way to generate both hidden and input decay rate based on the dynamics of the data. Experiment on Physio net dataset demonstrate that the proposed models outperformed existing GRU-D model in time series classification task. These experiments show that ODE based neural network model can successfully solve the informative missingness. The main goal of this work is to use an ODE solver as a black box  ODE solver and compute the gradient for the optimizer, which is in charge of optimizing the parameters of the neural network model.

\section{State of Art}
\label{sec:state_of_art}

\begin{figure}[h]
  \centerline{\includegraphics[trim=29 25 30 30,width=0.95\columnwidth]{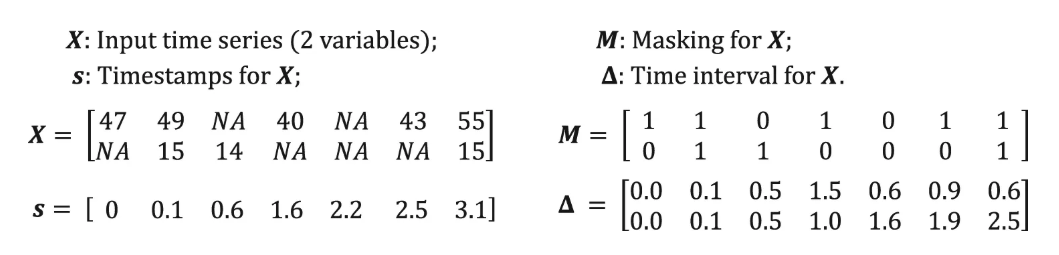}}
\caption{Continuous time series with missing values, from \cite{che2018recurrent}}\label{fig:fig4}
\end{figure}

A multivariate time series ($ X_{t} $) with $D$ variables  and of $T$ length  can be described as in \eqref{eq1}. Time series ($ X_{t} $) can have missing observation. Any observation for time series ($ X_{t} $) is $x_{t} \in \Re^{D}$ which represents $t$-th observation of all variables and $x_{t}^{d}$ represents the value of the $d$-th variable at time $t$.
\begin{equation}
\label{eq1}
X_{t} = {x_{1}, x_{2}, \ldots , x_{T}}^{D} \in \Re^{T \times D}
\end{equation}
The problem domain can be described as in Fig.~\ref{fig:fig4}. Here $X$ is a continuous multivariate time series with some missing values. The masking vector $M$ identifies missing observations. The timestamp vector $S$ records the time of the $t$-th observation of each variable.  The time interval vector $\Delta$ measures the duration of unavailability. The strength of the existing GRU-D implementation is that it uses the masking and time interval vectors to characterize the missing pattern rather than adopting the traditional missing-completely-at-random model. This approach helps quantify the influence of missing observations and adapt accordingly.

\section{Existing GRU-D implementation }

\label{sec:grud}

 The main contribution of GRU-D is that it can identify the long term dependencies as well as useful missing pattern in data. It can utilize long term temporal missing pattern in time series. Generally, GRU-D outperforms the other two RNN (GRU and LSTM) in terms of prediction experiments. The performance evaluation and comparison described in \cite{che2018recurrent} shows that GRU-D scored 2.5\% higher accuracy than other RNN implementation. Using GRU-D model, can predict of time series without relying on previous missing observations in the time series. This model achieves higher accuracy within less time in case of robust prediction of multivariate time series data. The core components of GRU-D are as following :

\subsubsection{Masking Vector} A masking vector (M)  $ m_{t} \in [0,1]^{D} $ indicates either the observation is available with value 1 or missing with value 0 at any time step $t$.

\begin{equation}
\label{eq3}
m_{t}^{d}= \begin{cases}
0 & \text{if $x_{t}^{d}$ is missing}
\\
1 & \text{if $x_{t}^{d}$ is available}
\end{cases}
\end{equation}

\subsubsection{Time Interval Vector} This vector computes the time interval $ {\delta_{t}}^{d} $ for each $d$-th variable since its last observation. If the first observation for $d$-th variable is measured at time $ s_{0} =0$, then the time interval can be measured by \eqref{eq4}.

\begin{equation}
\label{eq4}
\delta_{t}^{d} = \begin{cases}
s_{t} - s_{t-1} + \delta_{t-1}^{d} &  t>1, m_{t-1}^{d} =0
\\
s_{t} - s_{t-1} &  t>1, m_{t-1}^{d} =1
\\
0 & \text{otherwise}
\end{cases}
\end{equation}

\subsubsection{Data Imputation Value} For replacing the missing data, three different approaches, GRU-Mean, GRU-Froward and GRU-Simple, are used in the GRU-D model. 

\subsubsection*{GRU-Mean}: Each missing observation is replaced with the mean ($ x_{t}^{d} $) of the individual variable across the training examples. Both training and testing datasets is processed to compute the dynamics ($ \tilde{x}^{d} $) of GRU-Mean as shown in \eqref{eq:x_mean_1}. Here, $N$ is the length of time series.
\begin{equation}
\label{eq:x_mean}
x_{t}^{d} \leftarrow m_{t}^{d} x_{t}^{d}+\left(1-m_{t}^{d}\right) \tilde{x}^{d}
\end{equation}

\begin{equation}
\label{eq:x_mean_1}
\textrm{Where, } \tilde{x}^{d}=\sum_{n=1}^{N} \sum_{t=1}^{T_{n}} m_{t, n}^{d} x_{t, n}^{d} / \sum_{n=1}^{N} \sum_{t=1}^{T_{n}} m_{t, n}^{d}
\end{equation}

\subsubsection*{GRU-Forward}: Missing value is replaced with last available observations as shown in \eqref{eq:x_forward} . Here $t'<t$ is the last time when the $d$-th variable was observed.

\begin{equation}
\label{eq:x_forward}
x_{t}^{d} \leftarrow m_{t}^{d} x_{t}^{d}+\left(1-m_{t}^{d}\right) x_{t}^{d}
\end{equation}

\subsubsection*{GRU-Simple}: This approach identifies the missing variable along with the duration of missingness by concatenating the measurement, masking and time interval vectors as shown in \eqref{eq:x_simple}.

\begin{equation}
\label{eq:x_simple}
x_{t}^{(n)} \leftarrow\left[x_{t}^{(n)} ; m_{t}^{(n)} ; \delta_{t}^{(n)}\right]
\end{equation}

\subsubsection{ Hidden State Decay} 
\begin{figure}[htb]
 \centering
  \begin{subfigure}[b]{0.4\columnwidth}
    \includegraphics[width=\textwidth]{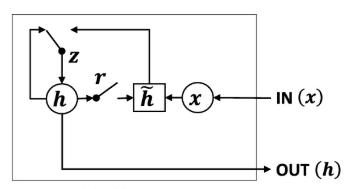}
    \caption{GRU Model Cell}
    \label{fig:gru}
  \end{subfigure}
  \hfill
  \begin{subfigure}[b]{0.58\columnwidth}
    \includegraphics[width=\textwidth]{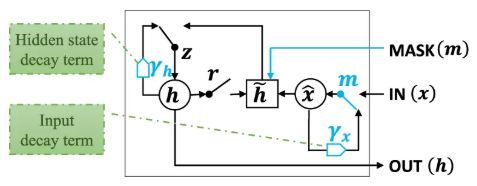}
    \caption{GRU-D Model Cell \cite{che2018recurrent}}
	\label{fig:grud}
  \end{subfigure}
  \caption{GRU vs GRU-D Model Cells}
\end{figure}

The importance of decay rate is that it can control the decay mechanism of the model considering underlying properties associated with each variable. As a result, individual variable of a multivariate time series can influence the prediction based on its actual weight of decay. The decay rate ($ \gamma_{t} $) is modelled as \eqref{eq:decay} with its associated weight and bias parameter $W_{\gamma}$ and $ b_{\gamma} $. The negative rectifier force decay rate to decrease monotonically between range [0,1].

\begin{equation}
\label{eq:decay}
\gamma_{t}=\exp \left\{-\max \left(0, W_{\gamma} \delta_{t}+b_{\gamma}\right)\right\}
\end{equation}
There are two different decay rate, hidden decay rate ($\gamma_{h_{t}}$) and input decay rate ($\gamma_{x_{t}}$). Hidden decay rate controls the decrease in decays of the previous hidden state $ h_{t-1} $ before computing current hidden state $ h_{t} $ as like \eqref{eq9}. Here, $\hat{h}_{t-1}$ represents a state between observations, but it has less control over the raw input variable. Therefore, a second decay rate, input decay rate, controls the decay of raw input variables directly.
\begin{equation}
\label{eq9}
\hat{h}_{t-1} = \gamma_{h_{t}} \odot h_{t-1}
\end{equation}
As shown in Figs.~\ref{fig:gru} and \ref{fig:grud}, the main difference between traditional GRU and GRU-D is that it uses two different decay mechanism hidden state and raw input. Using additional hidden state decay helps to capture the complete missing patterns.

The update functions for different gate units and intermediate states of GRU-D are explained by \eqref{eq:rt} (Reset Gate), \eqref{eq:zt} (Update Gate), \eqref{eq:tlilde_ht} (Intermediate state), and \eqref{eq:grud_ht} (Hidden state). The typical GRU cell update functions \cite{cho2014learning} are modified in GRU-D. The input ($ x_{t} $) and hidden ($ h_{t} $) vectors are replaced by $ \hat{y}_{t} $ and $ \hat{h}_{t} $ respectively. Besides, a new set of parameter vectors, $V_{r}, V_{z}$ and $ V $ are introduced for mask vector $ m_{t} $.

\begin{subequations}
  \begin{align}
    \label{eq:rt}
    r_{t} &= \sigma (W_{r}\hat{x}_{t} + U_{r}\hat{h}_{t-1} + V_{r}m_{t} + b_{r})
    \\
    \label{eq:zt}
    z_{t} &= \sigma (W_{z}\hat{x}_{t} + U_{z}\hat{h}_{t-1} + V_{z}m_{t} + b_{z})
    \\
    \label{eq:tlilde_ht}
    \tilde{h}_{t} &= \tanh(W\hat{x}_{t} + U(r_{t} \odot \hat{h}_{t-1}) + Vm_{t} + b)
    \\
    \label{eq:grud_ht}
    h_{t} &= (1- z_{t}) \odot \hat{h}_{t-1} + z_{t} \odot \tilde{h}_{t}
  \end{align}
\end{subequations}

\begin{figure}[h]
\centerline{\includegraphics[width=0.75\columnwidth]{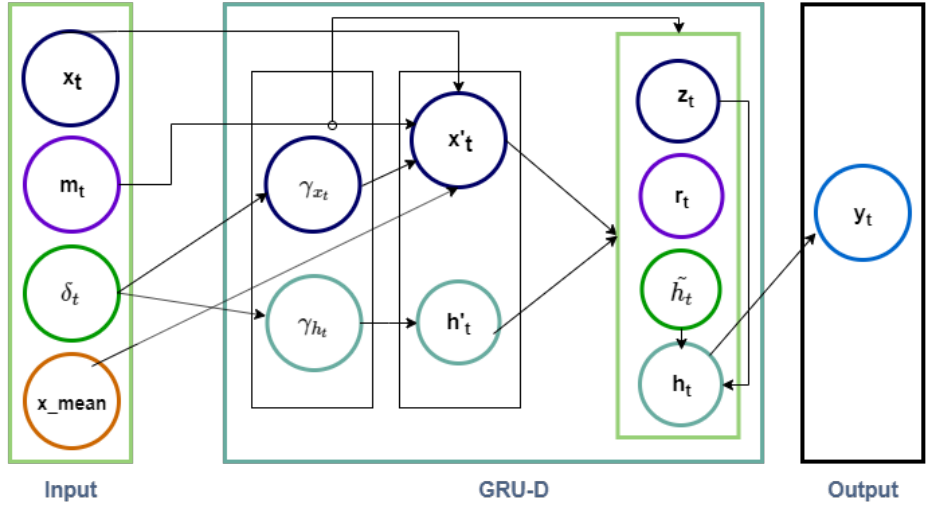}}
\caption{Architecture of GRU-D}\label{fig:fig3}
\end{figure}

The neural network architecture of GRU-D shown in Fig.~\ref{fig:fig3} shows that the input variable, masking vector and time interval vector is used as input for each cell in a GRU-D model. In Fig.~\ref{fig:fig3}, the GRU-Mean data imputation method is used. Both decay rates ($\gamma_{h_{t}}$ and $\gamma_{x_{t}}$) shape the input and hidden state.

\subsection{ODE-RNN}
\begin{figure}[h]
\centerline{\includegraphics[width=0.75\columnwidth]{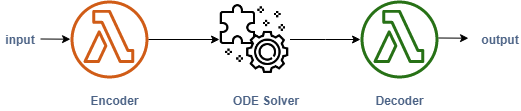}}
\caption{Architecture of ODE-RNN}\label{fig:fig8}
\end{figure}

 This neural network defines the hidden state between observations as the solution of an Ordinary Differential Equation (ODE) as shown in \eqref{eq:h-t}
\begin{subequations}
\begin{align}
\label{eq:h-t}
{h}'_{t}&= \ODESOLVE(f_{\theta}, h_{0},(t-1, t) )
\\
\label{eq:ht1}
h_{t}&= RNNCell({h}'_{t}, x_{t})
\end{align}
\end{subequations}
Here the function $f_{\theta}$  describes the dynamics of the hidden state using a neural network with parameters $\theta$. The previous hidden state at time $t-1$ can be computed at any time using a differential equation solver. ODE-RNN uses a standard RNN update function as shown in \eqref{eq:ht1} to compute current hidden state with input the intermediate state $h'_{t}$. The hidden state can be computed as \eqref{eq:ht1} without depending on the time interval implicitly. The hidden state is defined by an ODE solver and later updated by another RNN network at each observation. ODE-RNN \cite{rubanova2019latent} is an auto-regressive model which makes one step ahead prediction conditioned on the sequence of previous observation.  This model consists of two different neural networks: (i)~an encoder, (ii)~a decoder and ODE solver, as shown in Fig.~\ref{fig:fig8}.
\subsection{Latent-ODE}
latent-ODE is a variational auto-encoder based continuous time model. Both training and prediction use this auto-encoder. ODE-RNN is the encoder for Latent-ODE. This model can be characterized as encoder-decoder or fully ODE based sequence-to-sequence based architecture. Initial value problem is used to compute the latent state $( z_{t} )$ at any time $t$. A neural network (g) generates the final hidden state of ODE-RNN encoder as the mean and stranded deviation initial latent state $ z_{0} $ as shown in \eqref{eq:ode-rnn}. To get the approximate posterior both of the encoder and decoder are trained jointly. The posterior distribution over latent states are a function of final hidden state as shown in \eqref{eq:qz0}. This indicates the explicit uncertainty which is not available in traditional RNN or ODE-RNNs.

\begin{subequations}
\begin{align}
\label{eq:ode-rnn}
\mu_{z_{0}} , \sigma_{z_{0}} &= g (\text{ODE-RNN}_{\phi}(\left \{ y_{i},t_{i} \right \}_{i=0}^N))
\\
\label{eq:qz0}
q(z_{0}|) &= \eta (\mu_{z_{0}} , \sigma_{z_{0}})
\end{align}
\end{subequations}

The major benefit of latent-ODE is the hidden dynamics of the system and distribution of observations get decoupled from each other. As a result, each of the hidden state at any time $t$ can be computed independently. In addition, the posterior distribution over latent states can measure uncertainty which is a unique feature for latent-ODE.

\section{Methodology}
Two different models are proposed in this paper to solve the informative missingness of multivariate continuous time series as shown in Fig.~\ref{fig:fig4}.

\subsection{Continuous GRU-D}
\label{sec:odegrud}
 This model leverages an ODE solver in order to compute the hidden dynamics of continuous time series at any time $t$. Similar to \cite{che2018recurrent}, the hidden decay rate ($\gamma_{h_{t}}$) and input decay rate ($\gamma_{x_{t}}$) are computed using \eqref{eq:decay}. These decay rates are used as parameters for the network. However, instead of computing the hidden state as a dynamical system of discrete sequential value, this proposed model uses the initial value problem to calculate the continuous dynamics using derivatives generated by ODE solver. Therefore, the value of $d$-th variable at time $t$ ($ y_{t}$) is computed using \eqref{eq:yt}. \eqref{eq:x_simple} shows that $y_{t}$ is vector concatenating the value of last available observation at time $t$ ($ x_{t}$ ), masking vector ($ m_{t} $) value and hidden state value ($ h_{t} $) at time $t$. The hidden dynamics $ y_{0}$ is the initial value of the vector, $y_{0}^{n} \leftarrow\left[x_{0}^{n} ; m_{0}^{n} ; h_{0}^{n}\right]$. Here $n$ is the number of variable and $ y_{0}$ is an $ n \times 3 $ vector.  The resulting vector $ y_{t}$ provides the value of hidden state at time $t$ and observation value of $n$ variables at time $t$, $ y_{t} \leftarrow\left[x_{t}^{n} ; m_{t}^{n} ; h_{t}^{n}\right]$.

\begin{equation}
\label{eq:yt}
(y_{1}, y_{2}, \ldots, y_{t}) = \ODESOLVE( \mathscr{N}, y_{0}, T)
\end{equation}

\begin{figure}[h]
\centerline{\includegraphics[width=0.75\columnwidth]{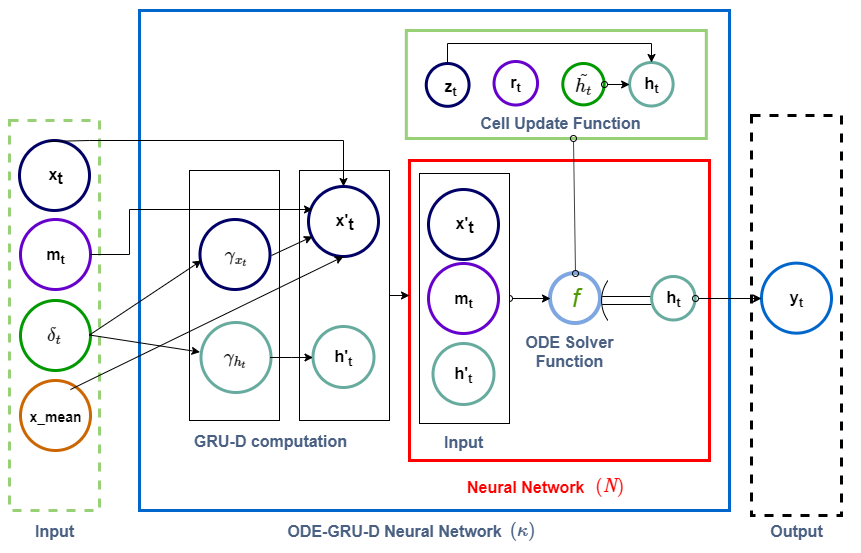}}
\caption{Architecture of ODE-GRU-D}\label{fig:fig6}
\end{figure}

Here $\mathscr{N}$ is a neural network similar to the update functions of existing GRU-D model cell, described in \eqref{eq:rt}, \eqref{eq:zt}, \eqref{eq:tlilde_ht}, and \eqref{eq:grud_ht}. An ODE neural network, as shown in Fig.~\ref{fig:fig6} is used to compute the update in hidden state and generate the target output.

This model leverage the ODE solver and uses the update functions as described in \eqref{eq:rt}, \eqref{eq:zt}, \eqref{eq:tlilde_ht} and \eqref{eq:grud_ht}. No additional training of encoder or decoder, unlike in \cite{chen2018neural} or \cite{rubanova2019latent}, is required in this proposed model. ODE-RNN \cite{rubanova2019latent} separately generates the hidden dynamics using a Neural ODE \cite{chen2018neural} and then computed hidden state is updated using standard RNN cell as shown in \eqref{eq:ht}. However,this proposed model continuously update the hidden dynamics using a Neural ODE network ($\mathscr{N}$) which use the update functions of GRU-D. \eqref{eq:ogrud}shows that the proposed neural network ($ \kappa $) takes the same inputs ($X, M, \Delta , T $) as like existing GRU-D as shown in Fig.~\ref{fig:fig4}. In addition to these inputs, the proposed model has  additional components (i) an ODE based neural network ($\mathscr{N}$) and (ii) a loss function ($ \mathscr{L} $) which helps the optimizer to leverage the dynamics of ODE in order to compute the update of hidden state. As shown in \eqref{eq:ogrud}, the time interval between two consecutive observations is not considered in hidden dynamics computation. Therefore, the length of time interval is irrelevant and the proposed model work for both higher as well as shorter sampling frequencies of any continuous time series.
\begin{equation}
\label{eq:ogrud}
y_{t} = \kappa (X, M, T, \mathscr{L},  \mathscr{N} )
\end{equation}

Algorithm~\ref{alg:ogrud} shows that a ODE based neural network \textit{(f)} updates the hidden state of the neural network. Here, \textit{(f)} is the update function that constructs the dynamics of hidden as well as derivative of the cell state and call the ODE solver to update all the parameters for optimizer at once. Algorithm~\ref{alg:updategrud} describes the steps for computing the derivative of hidden $dy$. The parameters of this neural network are optimized based on the output of loss function ($ \mathscr{L} $). The input for  $\gamma_{x_{t}}, \gamma_{h_{t}} $ are computed using the existing GRU-D computation described in \eqref{eq:decay}. Similarly, $x_{t}^{d}$ and $\tilde{h}_{t}$ are computed using  \eqref{eq:x_forward} and \eqref{eq9}. Here $\mathscr{N}$ is a neural network which uses $x_{t}^{d}$ and  $\hat{h}_{t}$ as input for a ODE based implementation of GRU-D cell update function. $T$ is the time series.

\begin{algorithm}[htb]
	\caption{Compute the state of an ODE-GRU-D cell}
	\begin{algorithmic}
		\Procedure{$\mathscr{N}$}{${x_{t}}^{d}, \tilde{h}_{t},T,\textit{hiddenDynamics}$}
		\State$y_{0}\gets \tuple(x_{0}^{d}, m_{0}, \hat{h}_{0})$
		\State$y_{t}, \frac{\delta{L}}{\delta{\theta}}\gets \ODESOLVE(f, y_{0}, T, h, \theta)$
		\State \textbf{return} $y_{t}$
		\EndProcedure
	\end{algorithmic}
\label{alg:ogrud}
\end{algorithm}

Algorithm~\ref{alg:updategrud} uses the same update functions as \cite{che2018recurrent} to compute the continuous hidden dynamics of the neural network.
\begin{algorithm} [htb]
	\caption{Update the derivative, the hidden state, and parameters for the optimizer}
	\begin{algorithmic}
		\Procedure{$\mathscr{N}$}{$y,T,\theta$}
		\State$x, h,m \gets y$
		\State Compute $r_{t}$ using \eqref{eq:rt}
		\State Compute $ z_{t} $ using \eqref{eq:zt}
		\State Compute $\tilde{h}_{t}$ using \eqref{eq:tlilde_ht}
		\State$h \gets (h - \tilde{h}_{t}) * z_{t}$
		\State$y_{t} \gets \tuple(x,h,m) $
		\State \textbf{return} $y_{t}$
		\EndProcedure
	\end{algorithmic}

\label{alg:updategrud}
\end{algorithm}

\subsection{Extended ODE-GRU-D}

\begin{figure}[h]
\centerline{\includegraphics[width=\columnwidth]{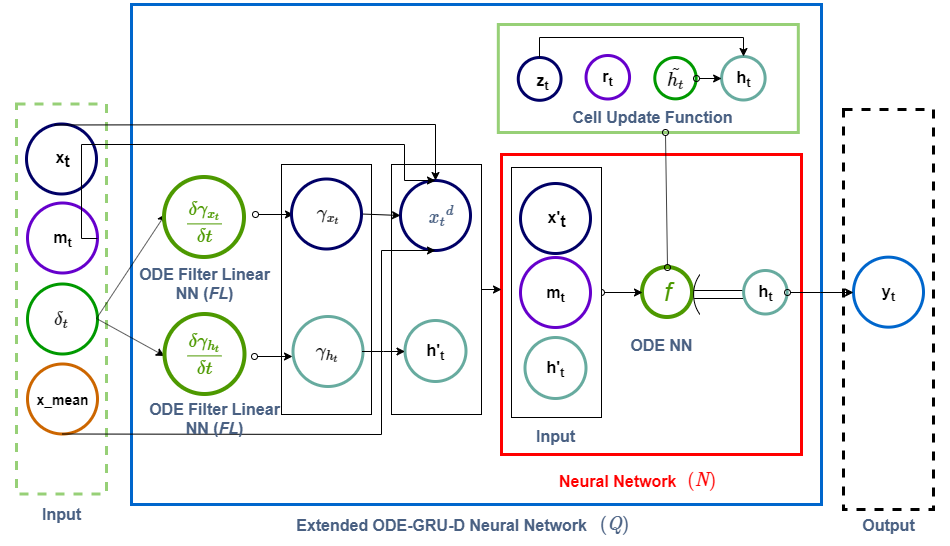}}
\caption{Architecture of Extended ODE-GRU-D Model}\label{fig:fig7}
\end{figure}

The second proposed model ($\mathscr{Q}$) characterized the missing pattern of the time series using Adjoint ODE solver. Fig.~\ref{fig:fig7} shows the architecture of proposed Extended ODE-GRU-D neural network model. Existing GRU-D model computes decay rate as shown in \eqref{eq:decay}, where $W_{\gamma}$ and $ b_{\gamma} $ are parameters of the same GRU model, and they are trained jointly with all the other parameters. However, if the dynamics of the decay rate can be computed as the derivative original input $\Delta$, with  respect to time, it can provide the accurate value of decay rate over time. Therefore, in this proposed model, a Filter Linear neural network $\textit{FL}$ is used to compute the derivative of decay rate over time. $W_{\gamma}$ and $ b_{\gamma}$ are parameters of $\textit{FL}$, and they are trained separately to compute the derivative of decay rate ($ \frac{\partial \gamma}{\partial t} $) as shown in \eqref{eq:ogrudext1}. ODE solver solves the neural network (\textit{FL}) with respect to $\Delta$ over time to compute the derivative. This derivative is used for computing the decay rate as shown in \eqref{eq:decay1}.

Fig.~\ref{fig:fig7} shows that two ODE Solver using Filter Linear neural network as function is used in proposed model to compute hidden state decay rate and input decay rate. The third ODE Solver compute the update in hidden state of the model.
\begin{subequations}
\begin{align}
\label{eq:ogrudext1}
\dfrac{\delta{\gamma}}{\delta{t}} &= \ODESOLVE (\textit{FL}, M,  h_{0}, T, \mathscr{L},  \mathscr{N} )
\\
\label{eq:decay1}
\gamma_{t}&=\exp \left\{-\max \left(0, \dfrac{\delta{\gamma}}{\delta{t}}\right)\right\}
\end{align}
\end{subequations}
This $ \gamma_{t} $ from \eqref{eq:decay1} compute  $ x_{t}^{d} \in  X^{D}$ using \eqref{eq:xtd}.

\begin{equation}
\label{eq:xtd}
x_{t}^{d} \leftarrow m_{t}^{d} x_{t}^{d}+\left( \gamma_{x_{t}^d} x_{t}^{d} + \left(  1-  \gamma_{x_{t}^d} \right) \right) x_{{\textit{mean}_{t}}}^{d}
\end{equation}
$ X^{D} = \{ x_{0}^{d},x_{1}^{d}, ...., x_{T}^{D} \}$  is used as the input for the proposed Extended ODE-GRU-D model. Similar to proposed ODE-GRU-D described in §\ref{sec:odegrud}, Extended ODE-GRU-D also uses ODE based neural network ($ \mathscr{N} $) to compute the update in hidden dynamics of the continuous time series as shown in \eqref{eq:ogrudext}. Here, $ h_{0} $  is the initial hidden state.
\begin{equation}
\label{eq:ogrudext}
y_{t} = \mathscr{Q} (X^{D}, M,  h_{0}, T, \mathscr{L},  \mathscr{N} )
\end{equation}

Proposed Extended ODE-GRU-D exploits ODE solver to understand the decay dynamics of the time series, therefore, controlling of change in the decay rate is more accurate in comparison to existing GRU-D model.this model learns the decay mechanism for raw input as well as the hidden dynamics of the continuous time series. The main advantages of this proposed models is that it naturally handle the decay mechanism and the gap between  consecutive observations.

\begin{figure*}[t!]
  \newlength{\sfl}
  \setlength{\sfl}{0.20\textwidth}
 \centering
  \begin{subfigure}[b]{\sfl}
    \includegraphics[width=\textwidth]{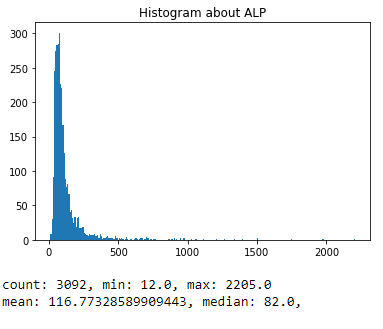}
    \caption{ALP}
    \label{fig:alp}
  \end{subfigure}
  \hfill
  \begin{subfigure}[b]{\sfl}
    \includegraphics[width=\textwidth]{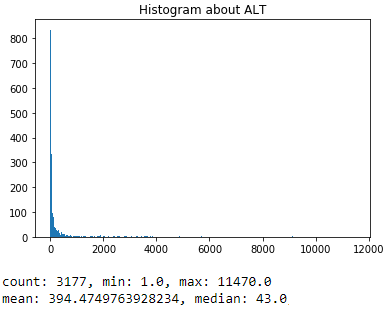}
    \caption{ALT}
	\label{fig:alt}
  \end{subfigure}
  \hfill
   \begin{subfigure}[b]{\sfl}
    \includegraphics[width=\textwidth]{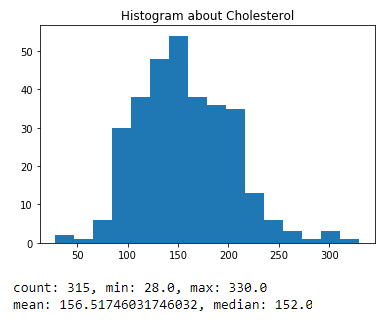}
    \caption{Cholesterol}
	\label{fig:cholestorol}
  \end{subfigure}
  \hfill
   \begin{subfigure}[b]{\sfl}
    \includegraphics[width=\textwidth]{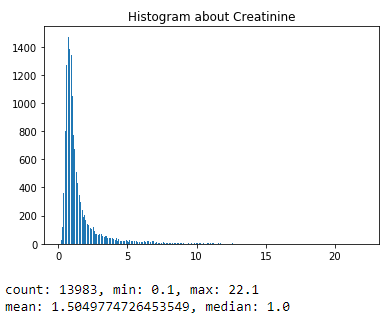}
    \caption{Creatinine}
	\label{fig:creatinine}
  \end{subfigure}
  \hfill
  \begin{subfigure}[b]{\sfl}
    \includegraphics[width=\textwidth]{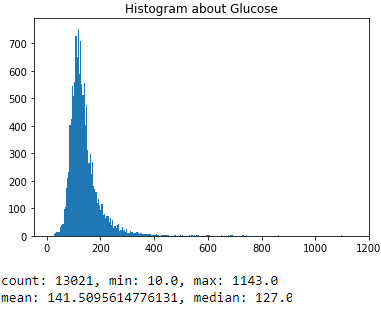}
    \caption{Glucose}
	\label{fig:glucose}
  \end{subfigure}
  \hfill
  \begin{subfigure}[b]{\sfl}
    \includegraphics[width=\textwidth]{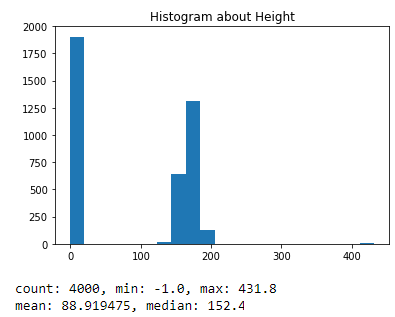}
    \caption{Height}
	\label{fig:height}
  \end{subfigure}
  \hfill
  \begin{subfigure}[b]{\sfl}
    \includegraphics[width=\textwidth]{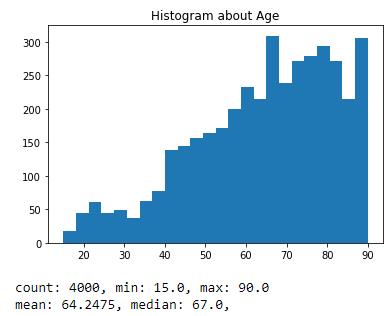}
    \caption{Age}
	\label{fig:age}
  \end{subfigure}
  \hfill
  \begin{subfigure}[b]{\sfl}
    \includegraphics[width=\textwidth]{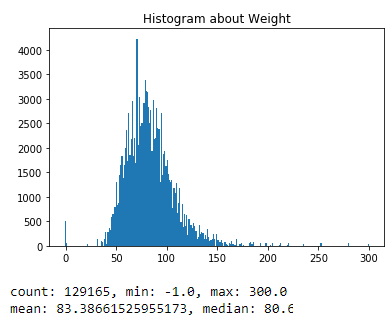}
    \caption{Weight}
	\label{fig:weight}
  \end{subfigure}
  \caption{Different parameters of the PhysioNet Challenges 2012 dataset}
  \label{fig:hist}
\end{figure*}

\section{Results}\label{AA}
As neural ODE is used to compute the hidden state at any time ($t$), no additional data processing is required for these tow proposed neural network models. Unlike ODE-RNN \cite{rubanova2019latent}, the proposed models uses single neural network. it does not require separate encoder and decoder.

\subsection{Dataset and Task Description}
For demonstrating the performance evaluation of proposed models, we use the PhysioNet Challenges dataset \cite{silva2012predicting}, a collection of multivariate time series with missing observations. It contains 8000 intensive care unit records (ICU), each of them a time series of 48 hours with 33 parameters. Fig.~\ref{fig:hist} shows the histogram for some of the parameters, i.e., ALP, glucose, age, height, weight, etc. This experiment uses Training Set~A for training as the output of this set is available only. As in the original paper, two prediction tasks, below, are attempted using this dataset.

\subsubsection*{Mortality task (Binary classification problem)} Predict patient's death in the hospital. There are 554 patients with positive mortality label.

\subsubsection*{Multi-task classification problem} Predict in-hospital mortality, length-of-stay less than 3 days,  patient's chance of having a cardiac condition, and patient' recovery state from surgery.

Table~\ref{tab:auc-table} shows that the proposed ODE-GRU-D and Extended ODE-GRU-D significantly outperform the existing GRU-D and Neural ODEs.For the comparative analysis area under curve (AUC) comparison.
\begin{table}[tb]
\centering
\caption{AUC on PhysioNet}
\label{tab:auc-table}
\normalsize
\begin{tabular}{lr@{$+$}l}
\toprule
\multicolumn{1}{c}{\textbf{Method}}
& \multicolumn{2}{c}{\textbf{AUC}} \\ \midrule
GRU-D & 0.824 &0.0.12 \\
ODE-RNN & 0.833&0.009 \\
Latent-ODE (RNN Encoder) & 0.781&0.018 \\
Latent-ODE (ODE Encoder) & 0.829&0.004 \\
Latent-ODE + Poisson & 0.826&0.007 \\
ODE-GRU-D & 0.8947&0.001 \\
Extended ODE-GRU-D & 0.9147&0.005 \\
\bottomrule
\end{tabular}
\end{table}

\section{Conclusions}

Most available solutions for handling missing value, in continuous time series, use data imputation. However, data imputation requires an extensive amount of data pre-processing; it also can impair the performance. This proposed model learns the hidden dynamics of time series as progress. It can compute the hidden state at any time $t$ without directly depending on the previous value. As the proposed model learns the derivative of changes it does not depends on additional data imputation. The derivatives of hidden state help to differentiate between longer and shorter dependencies; therefore, the proposed model works evenly fine for data with a higher sampling rate as well as a shorter sampling rate. As for using ODE based neural network, this proposed model requires more training time in comparison to testing time. Training time is relative to the dataset. Also, based on implementation, memory cost can be either fixed or incremental. Future extension of this work mainly focuses on using a similar model when the informative missingness in any dataset is not random. Instead, it follows a particular distribution. If the time pattern of occurrence of the missing observation can be identified, it is possible to switch on or off the computation of derivative. That can help to predict the missingness of the information and design the model accordingly.

\bibliographystyle{unsrt}
\bibliography{references}  


\end{document}